\documentclass[conference]{IEEEtran}
\usepackage{times}

\usepackage[numbers]{natbib}
\usepackage{multicol}
\usepackage[bookmarks=true]{hyperref}
\usepackage{algorithm} 
\usepackage{algpseudocode} 
\usepackage{amsmath,amssymb,amsfonts}
\usepackage{graphicx}
\usepackage{caption}
\usepackage{afterpage}

\usepackage{booktabs} 
\usepackage{multirow}
\usepackage{makecell}

\newcommand{\Hquad}{\hspace{0.5em}}

\begin{document}

\title{Learning Long-Horizon Robot Manipulation Skills via Privileged Action}

\author{Xiaofeng Mao$^{1}$, Yucheng Xu$^{1}$, Zhaole Sun$^{1}$, Elle Miller$^{1}$, Daniel Layeghi$^{1}$, Michael Mistry$^{1}$

\\
$^1$University of Edinburgh
}

\twocolumn[{%
\renewcommand\twocolumn[1][]{#1}%
\maketitle
\includegraphics[width=\linewidth]{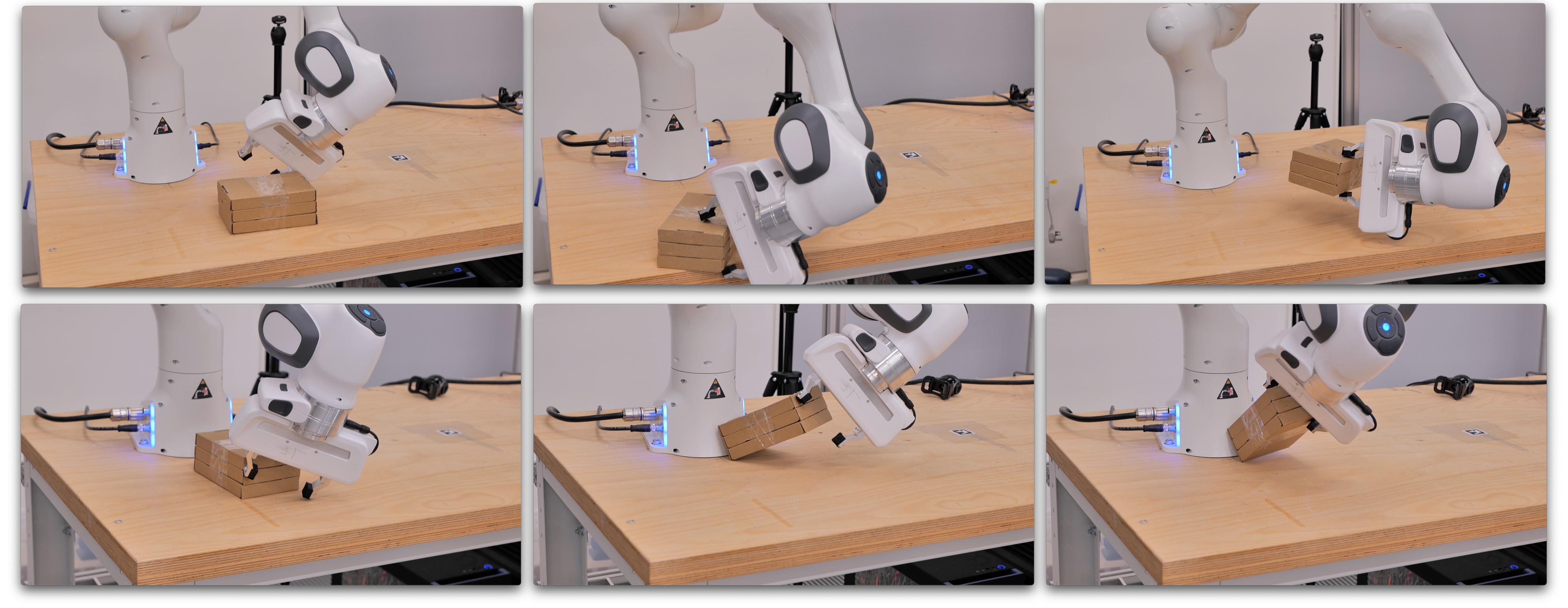}
\captionof{figure}{
Long-horizon manipulation incorporate distinct non-prehensile skills: push-and-grasp (upper) and pivot grasp (blow). These skills are learned across diverse environmental settings with our proposed framework while employing same reward function without any task-specific reward engineering.
\vspace{1.5em}
\label{fig:main}}
}]

\begin{abstract}
Long-horizon contact-rich tasks are challenging to learn with reinforcement learning, due to ineffective exploration of high-dimensional state spaces with sparse rewards.
The learning process often gets stuck in local optimum and demands task-specific reward fine-tuning for complex scenarios. 
In this work, we propose a structured framework that leverages privileged actions with curriculum learning, enabling the policy to efficiently acquire long-horizon skills without relying on extensive reward engineering or reference trajectories. Specifically, we use privileged actions in simulation with a general training procedure that would be infeasible to implement in real-world scenarios. These privileges include relaxed constraints and virtual forces that enhance interaction and exploration with objects.
Our results successfully achieve complex multi-stage long-horizon tasks that naturally combine non-prehensile manipulation with grasping to lift objects from non-graspable poses. 
We demonstrate generality by maintaining a parsimonious reward structure and showing convergence to diverse and robust behaviors across various environments.
Additionally, real-world experiments further confirm that the skills acquired using our approach are transferable to real-world environments, exhibiting robust and intricate performance. Our approach outperforms state-of-the-art methods in these tasks, converging to solutions where others fail.

\end{abstract}

\IEEEpeerreviewmaketitle

\section{Introduction}
Training robots for long-horizon, contact-rich manipulation tasks from scratch using reinforcement learning (RL) remains a substantial challenge. While RL has excelled in learning various complex locomotion and manipulation tasks~\cite{inhand, kalashnikov2022scaling, haarnoja2024learning}, most research focuses on optimizing specific short-horizon behaviors. The usual challenges associated with exploration compound dramatically with longer time horizons. If long-horizon sequences are considered, it is common to stitch multiple policies together~\cite{luo2024multi, mishra2023generative, ha2023scaling}. Overall this approach is undesirable, as we want the robot to autonomously discover optimal behaviors without handcrafting and combining primitives.

Exploration is yet again more challenging in contact-rich tasks~\cite{wang2024multi, ferrandis2024learning}. Imagine learning to grasp a cube floating in mid-air versus on a surface. The set of state-action pairs that leads to grasping in mid-air is significantly larger compared to on a surface. Physical collision boundaries can greatly impede exploration in high-reward regions, potentially trapping the learning process in local optimum. Furthermore, most contact-rich tasks involve learning to manipulate and change the state of external objects. The additional non-linear dynamics in these scenarios poses a greater challenge for learning.

To solve the exploration problem in long-horizon manipulation, recent work has explored using reference trajectories to provide a warm start and minimize unnecessary exploration~\cite{bauza2024demostart, chen2024object, triantafyllidis2023hybrid}. However, training robots from scratch remains a valuable area of research, especially when robots must operate in diverse environments where behavior varies significantly. Collecting demonstrations for each unique environment is labor-intensive and impractical at scale. Furthermore, limitations in teleoperation devices and mapping accuracy reduce the effectiveness of human teleoperation for teaching manipulation skills. When environmental parameters vary significantly the optimal policy can change substantially~\cite{yu2024skill, atanassov2024constrained}, and human-provided demonstrations may introduce biases that are misaligned with the optimal behavior of the robot due to differences in human and robotic capabilities.

Training robots in simulation offers interesting opportunities beyond what is feasible in the real-world. For example, simulations can easily provide privileged information, which has been shown to significantly improve sample efficiency and policy performance by providing useful features to learn from~\cite{jiang2023stable, miki2022learning, loquercio2021learning}. Despite this, privileged information is not guaranteed to help the robot discover innovative behaviors. We propose the novel concept of \textit{privileged actions}: actions that are infeasible in the real world, but enable efficient policy exploration. This includes relaxing constraints by disabling collisions and applying virtual forces to simplify interactions. We are motivated to solve long-horizon contact-rich tasks with RL, without requiring carefully tuned reward shaping. We introduce a framework that enhances exploration via privileged action, coupled with a learning curriculum to gradually align any physically infeasible training with real-world settings. The main contributions are summarised as follows:

\begin{itemize}
    \item \textbf{Novel concept of privileged action}:
    We propose to leverage privileged actions that are not feasible in the real world to simplify the problem and improve training efficiency.
    \item \textbf{General framework combing privileged action with curriculum learning}: 
    We build a general framework that enables the policy to efficiently solve long-horizon, contact-rich manipulation tasks while enforcing real-world constraints through a learning curriculum.
    \item \textbf{Robust Policy Adaptation}: Experiments demonstrate that the method adapts to environmental changes and converges to new behaviors without requiring reward modifications and with no reward indication on non-prehensile manipulation.
     Our approach outperforms state-of-the-art methods under identical setups, excelling across various tasks. Real-world experiments highlight the robustness of the learned behaviors.
\end{itemize}

\section{Related works}
\subsection{Robot Learning of Long-horizon Tasks}
Learning robotic policies for long-horizon manipulation tasks has been a longstanding and complex challenge. Many works focused on leveraging human prior knowledge to simplify this problem~\cite{cheng2023league, zhou2024spire, von2024art, wang2023mimicplay}.
Imitation learning is commonly used to simplify the complexities of long-horizon manipulation tasks. By breaking down long-horizon skills into sub-skills, imitation learning enables effective training of long-horizon policies, as shown in~\cite{luo2024multi, 10802807}. To further enhance learned behaviors, these policies can be improved through RL~\cite{triantafyllidis2023hybrid} or offline parameter optimization techniques~\cite{kumar2024practice}.

Training a long-horizon manipulation policy directly using RL often requires manual designed transitions between different primitives. To achieve robot grasping with external dexterity, \cite{zhou2023learning} incorporate a pre-generated grasp pose in the observation. They expanded the reward function to include the difference from the desired grasp pose and a penalty for collisions, to ensure the feasibility of the learned policy. Similarly, \cite{pmlr-v229-chen23e} split long-horizon tasks into a series of interconnected subtasks. They introduce a transition feasibility function that incrementally refines sub-policies to improve the success rate of chaining subtasks.
By splitting the non-prehensile manipulation into pre-contact and post-contact stages, \cite{kim2023pre} jointly train two policies where the pre-policy is used to determine the contact pose between the end-effector and the object, and the post-policy is used to apply action on the object. These two policies are jointly trained with a highly complex and fine-tuned reward function. In this work, rather than relying on human demonstrations or reward engineering to guide robots toward predefined optimal behaviors, we propose a framework that enables the policy to discover solutions autonomously. By simplifying the problem and expanding the state-action space through privileged actions, our approach addresses the inefficiency of RL in exploring sparse-reward environments.

\subsection{Curriculum Learning}
Curriculum strategies are widely used in RL to enable robots to master challenging tasks. These strategies naturally guide the learning process by starting with simpler tasks and gradually increasing complexity. For instance, \cite{CHIAPPA20243969} use a curriculum to enable a human hand model with 39 muscles to rotate two Baoding balls in its palm.

The work~\cite{chen2021system} devise a gravity-based curriculum to enable in-hand manipulation. They first train with the gravity vector pointing upwards, and then gradually decrease until the normal value. Additionally, \cite{bauza2024demostart} combine human demonstrations with an auto-curriculum strategy for dexterous manipulation. Demonstrations provide initial guidance to reduce the search space and accelerate policy convergence, while the auto-curriculum identifies areas requiring improvement and enhances them through RL. Similarly, \cite{margolis2024rapid} employ an adaptive curriculum based on velocity commands to train a robot to run and turn quickly on natural terrains. Our work leverages curriculum learning to gradually reduce the availability of privileged actions and guide the policy to a physically realistic solution.

\subsection{Privileged Information and Actions}
Simulations offer access to rich information that is often difficult to obtain in the real world.
Previous works have extensively leveraged simulation to provide privileged information to enable policies to acquire essential knowledge, resulting in robust performance~\cite{lee2020learning, chen2023visual, qi2023general, yang2024anyrotate}. Learning from privileged information improves policy learning by reducing the complexity of the state-action mapping required to be learned.
\cite{mordatch2012contact} employ a contact-invariant optimization (CIO) method to specify when and where contact should occur on an object, using hand movements to replace this auxiliary decision variable gradually. Likewise, \cite{cheng2023enhancing} utilized Monte Carlo Tree Search (MCTS) to explore contact points on objects, enabling robot manipulation with exceptional dexterity. 
By relaxing collisions between the robot and obstacles, \cite{zhuang2023robot} combined a curriculum strategy with a specifically designed reward with penetration terms to train a vision-based parkour policy.
In~\cite{zhou2023learning}, predefined grasp poses were used to learn grasping with extrinsic dexterity. Relaxed collisions and penalties for penetration were incorporated into the training process.

Despite the success of these methods, they often require either complex task specific formulation or finely crafted reward functions. To the best of our knowledge, we are the first to propose a structured framework for solving long-horizon manipulation tasks without introducing any delicately designed reward terms.

\section{Method}
Our method is a structured framework that provides a solution to solving a wide range of long-horizon manipulation tasks using privileged actions with curriculum learning. It does not rely on heavy reward shaping or human priors e.g. reference trajectories. As shown in Fig.~\ref{Fig:Figure_2}, the method follows a three-stage process consisting of constraint relaxation, virtual forces, and the normal setting. In this section, we first present the problem formulation of our work, then we introduce the three stages depicted in Fig.~\ref{Fig:Figure_2} in depth. After that, we introduce the reward settings we used to train our policy and how we conduct sim-to-real transfer by applying domain randomization and improved control bandwidth. In this work we focus specifically on manipulation, however our framework can be naturally extended to other robotic control tasks.
\begin{table}[h]
\centering
\renewcommand{\arraystretch}{1.3} 
\setlength{\tabcolsep}{10pt} 
\resizebox{0.98\columnwidth}{!}{
\begin{tabular}{|c|p{0.6\columnwidth}|} 
\hline
\textbf{Symbol} & \textbf{Definition} \\
\hline
$\mathbf{q}_t  = \begin{bmatrix}\mathbf{q}_{R,t}, \mathbf{q}_{O,t}\end{bmatrix} \in \mathbb{R}^{n_p = n_{p_{r}} + n_{p_{o}}}$ 
 & \textbf{System state}, representing the robot's and object's positions. \\[8pt]
 $\mathbf{\dot q}_t  = \begin{bmatrix}\mathbf{\dot q}_{R,t}, \mathbf{\dot q}_{O,t}\end{bmatrix} \mathbb{R}^{n_v =n_{v_{r}} + n_{v_{o}}}$ 
 & \textbf{System state}, representing the robot's and object's velocities. \\[8pt]
$\mathbf{x}_t  = \begin{bmatrix}\mathbf{q}_{R,t}, \mathbf{q}_{O,t}, \mathbf{\dot q}_{R,t}, \mathbf{\dot q}_{O,t}\end{bmatrix}\in \mathbb{R}^{n_p + n_v}$ 
 & \textbf{System state}, representing the positions and velocities of the robot and object. \\[8pt]

$\mathbf{u}_t 
 = \begin{bmatrix}\mathbf{u}_{R,t}, \mathbf{u}_{O,t}\end{bmatrix} 
 \in \mathbb{R}^m$
 & \textbf{Control input}, partitioned into:
   \(\mathbf{u}_{R,t} \in \mathbb{R}^{m_R}\) (robot joint torques/controls) and 
   \(\mathbf{u}_{O,t} \in \mathbb{R}^{m_O}\) (linear force applied to the object). \\[8pt]

$\mathbf{f}(\mathbf{x}_t)$ 
 & \textbf{Passive dynamics}, including inertial and external contact forces. \\[8pt]

$\mathbf{g}(\mathbf{x}_t)$
 & \textbf{Control influence} on state evolution. \\[8pt]

$\mathbf{B}(\mathbf{x}_t) \in \mathbb{R}^{n_{v_{o}} \times m_O}$
 & \textbf{Gating matrix} that regulates the force applied to the object. \\[8pt]

$\mathbf{M}(\mathbf{q}_t) \in \mathbb{R}^{n_v \times n_v}$
 & \textbf{Mass/inertia matrix} for the robot and object in generalized coordinates. \\[8pt]

$\mathbf{c}(\mathbf{x}_t)$ 
 & \textbf{Bias term} accounting for Coriolis, gravity, and frictional forces. \\[8pt]

$\mathbf{J}(\mathbf{q}_t)$
 & \textbf{Jacobian} mapping contact-space forces to joint torques. \\[8pt]

$r(\mathbf{x}_t, \mathbf{u}_t)$ 
 & \textbf{Reward function} at each timestep. \\[8pt]

$\gamma \in (0,1]$ 
 & \textbf{Discount factor} for long-horizon returns. \\[8pt]

$\delta_p, \delta_v > 0$ 
 & \textbf{Distance and velocity thresholds} for gating object forces. \\[8pt]

$\alpha$ 
 & \textbf{Curriculum factor} controlling $\delta_p$ and $\delta_v$ thresholds. \\[8pt]

$\phi_R(\mathbf{x}_t), \phi_O(\mathbf{x}_t)$ 
 & \textbf{Signed distances} to the table for the robot’s end-effector and the object. \\[8pt]

$\mathbf{F}_{R,t}, \mathbf{F}_{O,t}$ 
 & \textbf{Normal contact forces}, e.g., forces between the robot, object, and table. \\[8pt]

$\Delta_R$ 
 & \textbf{Penetration offset} regulating robot-table collision relaxation. \\[8pt]

$\mathbf{x}_0 \sim p_0$ 
 & \textbf{Initial state distribution}. \\
\hline
\end{tabular}
}
\caption{Symbol definitions used in our framework.}
\label{tab:symbols}
\end{table}

\subsection{Problem formulation}
Let us consider the model-free RL setting that corresponds to manipulating objects on a tabletop. In this case our state consists of the positions and velocities of a robot and an object that is to be manipulated $\mathbf {x} =\begin{bmatrix}\mathbf{q}_{R,t}, \mathbf{q}_{O,t}, \mathbf{\dot q}_{R,t}, \mathbf{\dot q}_{O,t}\end{bmatrix}$. The transition of this state is typically formulated as a Markov Decision Process (MDP). At each time step $\mathbf{t}$, the policy $\mathbf{\pi_{\theta}}$ predicts the action $\mathbf{u}$ based on the current observation $\mathbf{x}$.
The objective of training the policy $\pi_{\theta}$ is to maximise the discounted return over the episode length $T$. To perform this maximisation, RL frameworks typically use a simulator which internally solves an optimisation problem to compute physically realistic motions, accelerations and forces. Below, we present these elements into a concise mathematical program which provides an overview of the general tabletop manipulation problem with its parameters defined in table \ref{tab:symbols}.

\noindent\textbf{Maximize:}
\begin{equation} \label{eq:objective}
J(\mathbf\theta) 
= \mathbb{E}_{\mathbf{x}_0 \sim p_0,\, \mathbf{u}_t \sim \pi_\theta(\cdot \mid \mathbf{x}_t)} 
\Bigl[\sum_{t=0}^\infty \gamma^t\, r(\mathbf{x}_t, \mathbf{u}_t)\Bigr]
\end{equation}

\noindent\textbf{Subject to:}
\begin{align}
\mathbf{x}_{t+1} 
&= \mathbf{x}_t 
+ \Bigl[\mathbf{f}(\mathbf{x}_t) \;+\; \mathbf{g}(\mathbf{x}_t)\,\mathbf{u}_{t}\Bigr]\;\Delta t, 
\label{eq:dynamics} \\[5pt]
\mathbf{f}(\mathbf{x}_t)  
&= 
\begin{bmatrix}
\dot{\mathbf{q}}\\[4pt]
\mathbf{M}\bigl(\mathbf{q}\bigr)^{-1}
\Bigl(
  \mathbf{J}\bigl(\mathbf{q}\bigr)^\top 
  \,  \begin{bmatrix}
    \mathbf{F}_{R,t} \\
    \mathbf{F}_{O,t}
  \end{bmatrix}
  ~-~
  \mathbf{c}\bigl(\mathbf{x}_t\bigr)
\Bigr)
\end{bmatrix},
\label{eq:passive_dynamics} \\[5pt]
\mathbf{g}(\mathbf{x}_t)
&=\;
\begin{bmatrix}
\mathbf{0}\\[5pt]
\mathbf{M}\bigl(\mathbf{q}\bigr)^{-1}
\begin{bmatrix}
\mathbf{I}_{n_{v_r}\times m_R}\\
\mathbf{0}_{n_{v_o}\times m_R}
\end{bmatrix}
\end{bmatrix},
\label{eq:control_dynamics_prog} \\[6pt]
\mathbf{u}_{R,t}
&\sim \mathbf{\pi}_\mathbf{\theta}(\cdot \mid \mathbf{x}_t), 
\quad 
t = 0, 1, 2, \dots
\label{eq:policy}
\end{align}

Maximising the reward in this case means overcoming the challenges inherent to the environment. For example the state of the object can only be changed through contact forces applied by the robot. If the object is in a non-graspable pose, it becomes particularly difficult for the robot to determine how to interact with object from scratch under sparse rewards. Equation \ref{eq:control_dynamics_prog} formally describes this constraint showing that any direct forces on the object via the policy are nullified. This challenge arises due to the involvement of non-prehensile manipulation to change the object to a grasp pose, which is inherently contact-rich and requires extensive long-horizon exploration at this stage.

Privileged actions serve as an effective technique to simplify this problem, facilitating more efficient exploration during policy learning. By reducing collision complexity and minimizing the need for direct interaction with the object, the robot can accomplish the task more easily, guiding the robot state-action space toward a more feasible subset. This strategy mitigates the risk of the robot becoming trapped in local optimum and significantly improves its exploration capability. The following subsections present a detailed discussion of privileged actions, reward setting, and the auto-curriculum framework.

\subsection{Constraint Relaxation with Collision Management}

\begin{figure*}[t]
\centering
\includegraphics[width=1.0\linewidth,keepaspectratio]{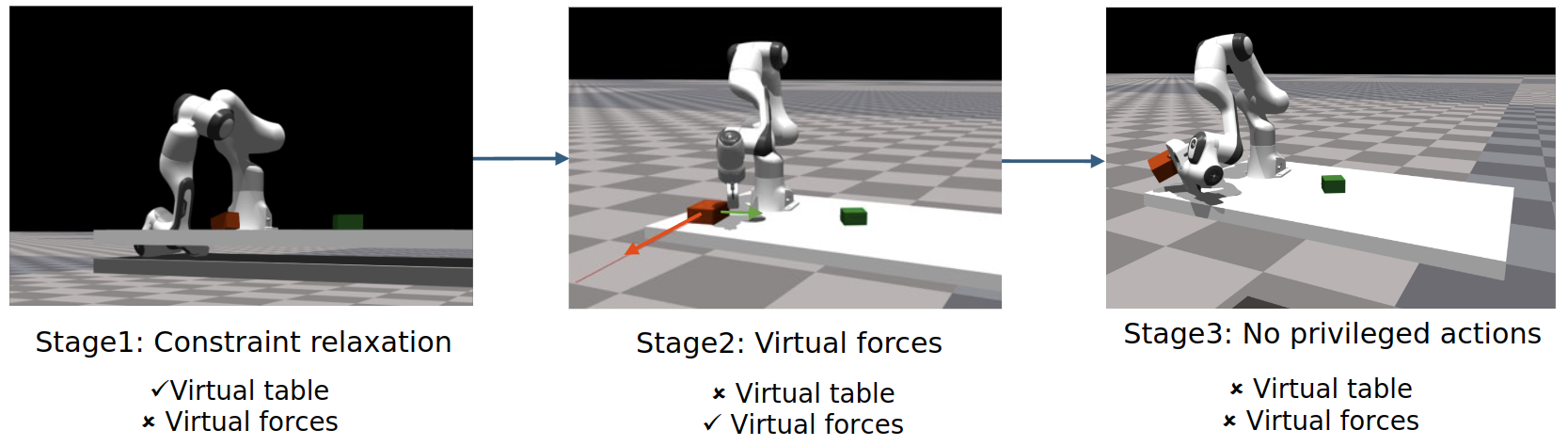}
\caption{
A structured framework utilizes privileged actions with curriculum learning. In stage 1, the robot penetrates the white table while a grey virtual table limits this penetration and is gradually lifted during training; stage 2 applies virtual forces on the object, indicated by blue and green arrows, and by stage 3, no privileged actions is used.
}
\vspace{2mm}
\label{Fig:Figure_2}
\end{figure*}

When grasping an object from an initially ungraspable pose, the table surface may be considered as an obstacle that impedes the robot from achieving a successful grasp. 
We first train the policy with constraint relaxation, by cancelling the collision between the robot and the table. This allows the robot to learn the manipulation skill more effectively. 
Concretely, we make the contact constraint forces between the robot and the table $\mathbf{F}_{R,t}$ less restrictive by increasing the distance at which contact is triggered, $\phi_R(\mathbf{x}_t)$ by $\Delta_R$.
\begin{equation}
\mathbf{F}_{R,t} \geq \mathbf{0}, 
\Hquad
\phi_R(\mathbf{x}_t) + \Delta_R \;\geq 0, 
\Hquad
\bigl(\phi_R(\mathbf{x}_t) + \Delta_R\bigr)\,\mathbf{F}_{R,t} = 0, 
\label{eq:contact_complementarity}
\end{equation}
However, it is important to note that constraint relaxation expands the robot's state-action space, potentially causing significant deviations in the action distribution. For instance, the robot may learn to lift the object using its arm rather than the gripper, leading to incorrect behaviors that create challenges in later training stages. To mitigate this issue, we introduce a virtual table that interacts with the robot and gradually increase its height until it aligns with the actual table surface. This process is illustrated in the left figure of Fig.~\ref{Fig:Figure_2}, where the white table does not collide with the robot, while the grey table represents the virtual surface that enforces collision constraints. In this work, we initially set $\Delta_R = [0, 0.3]$ (i.e.\ 30cm below the actual surface). As the policy meets the success rate condition, the virtual table is progressively raised.

\subsection{Virtual force}
When restoring the collision relationship between the robot and table, the robot follows the previously learned policy, often attempting penetration actions. It is important to highlight that the constraint relaxation employed in~\cite{zhuang2023robot, zhou2023learning} is effective primarily due to dense rewards that penalise penetration.When transitioning back to real collision dynamics, the dense rewards ensures that the robot receives timely reward feedback for state changes, allowing it to quickly adapt to the updated scenario.

Learning the interaction between the robot and the object presents a significant challenge, as the subset of the state-action space capable of inducing meaningful object state changes is much narrower than the robot’s original state-action space. The robot must establish contact with the object to induce state transitions and maintain this contact to achieve effective manipulation. Consequently, training an RL policy for non-prehensile manipulation without reward guidance is highly challenging. To overcome this difficulty, we introduce a virtual force to promote the interaction between the robot hand and the object. Specifically, we design the trained policy to predict the force applied to the object while enforcing a constraint that ensures the robot's end-effector actions align with the virtual force. 
Formally, the control is sampled from  $[\mathbf{u}_{R,t},
\mathbf{u}_{O,t}]
\sim \mathbf{\pi}_\mathbf{\theta}(\cdot \mid \mathbf{x}_t)$ additionally the space in which the control can act on is modified to
\begin{align}
\mathbf{g}(\mathbf{x}_t)
&=\;
\begin{bmatrix}
\mathbf{0}\\[5pt]
\mathbf{M}\bigl(\mathbf{q}\bigr)^{-1}
\begin{bmatrix}
\mathbf{I}_{{n}_{v_r}\times m_R} & \mathbf{0}\\
\mathbf{0} & \mathbf{B}\bigl(\mathbf{x}_t\bigr)
\end{bmatrix}
\end{bmatrix},
\label{eq:control_dynamics} \\[6pt]
\mathbf{B}(\mathbf{x}_t) 
&= 
\begin{cases}
\mathbf{I}_{n_{v_o} \times m_O}, & 
\text{if } 
 \bigl\|\mathbf{q}_{O,t} - \mathbf{q}_{EE,t}\bigr\| < \delta_p \cdot \alpha \\ 
& \;\wedge\; 
\bigl\|\dot{\mathbf{q}}_{O,t} - \dot{\mathbf{q}}_{EE,t}\bigr\| < \delta_v \cdot \alpha,\\
\mathbf{0}_{n_{v_o} \times m_O}, & \text{otherwise}.
\end{cases} 
\label{eq:gating_matrix}\nonumber \\[6pt]
\end{align}
The gating matrix $\mathbf{B}(\mathbf{x_t})$ enables control action forces on the object. Specifically, $\mathbf{B}(\mathbf{x}_t)$ activates $\mathbf{I}$ only if 
$\bigl\|\mathbf{q}_{O,t} - \mathbf{q}_{EE,t}\bigr\| < \delta_p\,\alpha$ 
and 
$\bigl\|\dot{\mathbf{q}}_{O,t} - \dot{\mathbf{q}}_{EE,t}\bigr\| < \delta_v\,\alpha$, 
ensuring that the policy’s force $\mathbf{u}_{O,t}$ affects the object only when the end-effector is sufficiently close in position and velocity. Otherwise, $\mathbf{B}(\mathbf{x}_t)$ is zero, leaving the object unactuated. 
Overall, the curriculum learning is conducted to encourage the robot movement to gradually replace and approximate the virtual force.  

In this work, we initially set $\delta_p = 10$, $\delta_v = 5$, and $\alpha = 0.85$, which means that there are no restrictions during the early stage of training. Additionally, the virtual force applied on the object is restricted to the $\textbf{x}$ and $\textbf{y}$ directions. Once the environment achieves success, these thresholds are progressively reduced by multiplying them with a factor $\alpha \in [0.06, 0.85]$.

\begin{equation}
\alpha =
\begin{cases}
    \text{clamp}(\alpha*0.9, 0.06, 0.85) & \text{if reach success} \\
    \alpha & \text{otherwise.}
\end{cases}
\end{equation}

\subsection{Reward Setting}
Dense rewards are often used to guide learning.
However, relying on heavily shaped reward functions offers limited generalisability and can demand substantial human effort.

Thus, we adopt the default reward for grasping and lifting objects from the table, as provided by IsaacGym~\cite{makoviychuk2021isaac}, without introducing any additional modifications.. Specifically, the task we focus on in this work involves non-prehensile manipulation with multi-stage behavior. The reward for non-prehensile manipulation is sparse, as no explicit rewards are provided to guide this behavior. The reward we used in this task is defined as follows:

\begin{equation}
r_{total} = r_{f} + r_{l} + r_{k} + r_{p} + r_{b}
\end{equation}

Where $r_{f}$ is the distance reward between the robot end-effector and the object, $r_{l}$ is the lifting reward, and $r_{k}$ is the distance reward between the object and the goal. $r_{p}$ is the penalty term, which reduces jerk motion by penalizing sudden changes in movement. Additionally, $r_{b}$ is the bonus reward for successfully reaching the goal.

Previous works utilising relaxed collision constraints added extra reward terms to guide the behavior of the robot~\cite{zhuang2023robot, zhou2023learning}. However, such methods require careful fine-tuning of the reward function; otherwise, the local optimum of the reward function may shift, negatively impacting learning. 
In this work, we do not incorporate any additional rewards to guide the robot’s behavior when using privileged actions. Instead, we employ an auto-curriculum framework that allows the policy to learn autonomously and efficiently, gradually transitioning from privileged actions to the standard setting through a curriculum strategy.

\subsection{Structured Privilege Actions with Curriculum Learning}
The curriculum learning strategy is an effective approach for training robots to master complex tasks by initially focusing on simpler ones. In this work, we employ a curriculum strategy to guide the policy in progressively learning applicable behaviors by gradually reducing privileged actions. The detailed process is presented in Algorithm 1.

\begin{algorithm}
    \caption{Three-Stage Curriculum Training} 
    \begin{algorithmic}[1]
        \State \textbf{Initialize:} $\Delta_R \gets 0.3$, $\alpha \gets 0.85$, $\text{epoch} \gets 0$
        \State \textbf{Set:} $\Delta_{i} \gets 0.1$, $\alpha_d \gets 0.9$, $\alpha_{min} \gets 0.06$
        \While{not end of training}
            \If{$\Delta_R > 0.0$} \Comment{Stage 1: Virtual surface}
                \State Train policy $\pi_\theta$ with constraint relaxation
                \If{success rate $> 70\%$}
                    \State $\Delta_R \gets \Delta_R - \Delta_{i}$
                \EndIf
            \ElsIf{$\alpha > \alpha_{min}$} \Comment{Stage 2: Virtual force}
                \State Train policy $\pi_\theta$ with virtual force
                \If{success rate $> 70\%$}
                    \State $\alpha \gets \max(\alpha \cdot \alpha_d, \alpha_{min})$
                \EndIf
            \Else \Comment{Stage 3: No privileged actions}
                \State Train policy $\pi_\theta$  without privileged actions
            \EndIf
        \EndWhile
    \end{algorithmic} 
\end{algorithm}

\begin{figure}[t]
\centering
\includegraphics[width=0.8\linewidth,keepaspectratio]{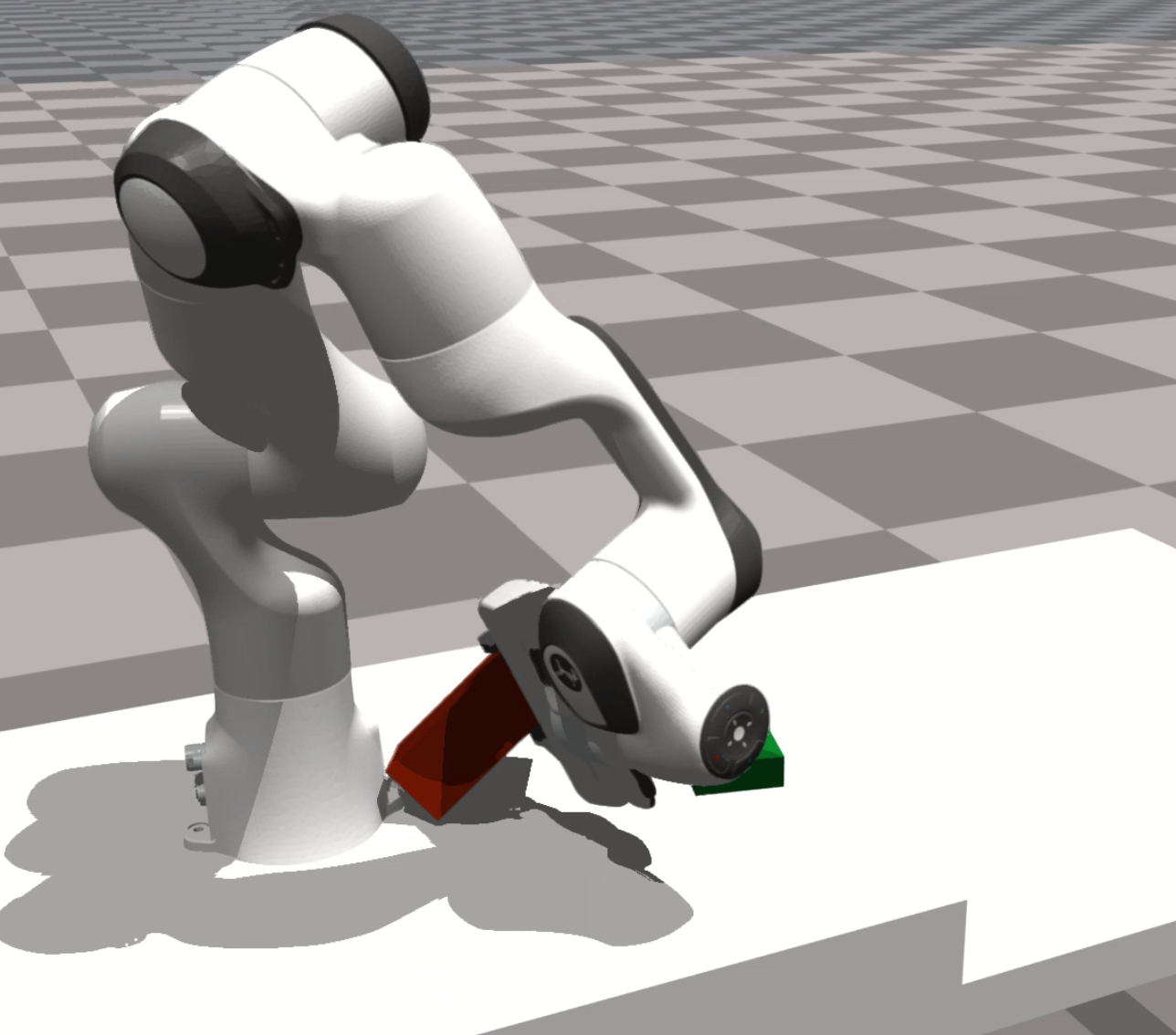}
\caption{
Franka lift cube from a non-graspable pose in a constrained environment, where small walls placed at the table edge block direct pushing behavior. The policy learns to use the base of the robot as support to execute a pivot grasp.}
\label{Fig:Figure_1}
\vspace{-5mm}
\end{figure}

\section{Experiment Result}
In this work, we focus on long-horizon, contact-rich tasks involving non-prehensile manipulation combined with grasping to lift objects from non-graspable poses. All tasks are set up and trained using the IsaacGym simulator~\cite{makoviychuk2021isaac}.

We conduct two tasks to validate the performance of our proposed approach. The first task involves grasp and lift object from non-graspable pose using a Franka robot arm and a Franka gripper. The second task, performed in simulation, involves grasping and in-hand manipulation of several YCB objects~\cite{calli2017yale}, which is a more challenging task due to the use of a dexterous hand with higher degree of freedom. These experiment demonstrates the generality of our method across different setups.

Additionally, we conduct real-world experiments to show that our approach not only enables efficient policy convergence to robust solutions in simulation but is also adaptable across various platforms and tasks. 
More importantly, when the environment changes, with the same reward and did not include any further guidance on the non-prehensile manipulation behavior, our method still allows the policy to adapt and converge to physically acceptable and task-specific behaviors.

\subsection{Grasp and lift object from non-graspable pose}\label{Franka_normal}
\textbf{Experiment setup.} 
We evaluate our method using the Franka robot to grasp an object in an ungraspable pose. The Franka robot is set up on a table. The object used in the experiment measures 15cm $\times$ 10cm $\times$ 6cm, while the maximum opening distance of the Franka gripper is 8cm. When the object is lying flat on the table, the Franka gripper cannot directly grasp the object due to the limited opening distance of its gripper. Instead, the robot must learn how to push the object to the edge of the table and grasp it from the object's side. 

We further evaluated our method by placing the Franka robot in a more constrained environment with small walls surrounding the table. This setup prevented the robot from using its previous strategy of pushing the object to the edge for side grasping. 
The observation and reward used in this task is the same as in~\ref{Franka_normal}.
Despite these constraints, our method enabled the robot to develop a stable and effective behavior, leveraging its own frame as support to do the pivot grasp. Traditionally, such behaviors require carefully designed rewards or human guidance, but our approach eliminates the need for manual intervention.

Additionally, we tested both tasks with the Franka robot in real-world scenarios, demonstrating that the behaviors emerging from our method are not only robust but also physically valid and applicable in real-world environments.

\textbf{Results in simulation.} 
With our method, the Franka robot successfully demonstrated long-horizon behavior by pushing the object to the edge of the table, grasping it from the side, and lifting it. The policy also can adapt to various object pose, when the pose of the object is not suitable for directly push and grasp from the side, it will reorient the object first, and then grasp it. 
In the more constrained environment, where the original solution was blocked by walls, the policy also can adapt to it and leveraging the robot base as support to achieve a pivot grasp, all without human guidance in either the observation or reward function. 
In comparison, the vanilla PPO failed on both tasks. The robot end-effector remains at the center of the object and fails to lift, as this behavior is the local optimum of the reward function that minimize object to end-effector distance.
\begin{table}[h]
    \centering
    \renewcommand{\arraystretch}{1.2} 
    \setlength{\tabcolsep}{8pt} 
    \begin{tabular}{lcc}
        \toprule
        \textbf{Task} & \textbf{Our method} & \textbf{PPO} \\
        \midrule
        \textbf{Push and Grasp} & $(2.34 \pm 0.13)\times 10^4$ & $65\pm10$ \\
        
        \textbf{Pivot Grasp} & $(2.17 \pm 0.29)\times 10^4$ & $52\pm17$ \\
        
        \bottomrule
    \end{tabular}
    \caption{Reward comparison on different tasks.}
    \label{tab:results}
\end{table}

\begin{figure}[t]
\centering
\includegraphics[width=1.0\linewidth,keepaspectratio]{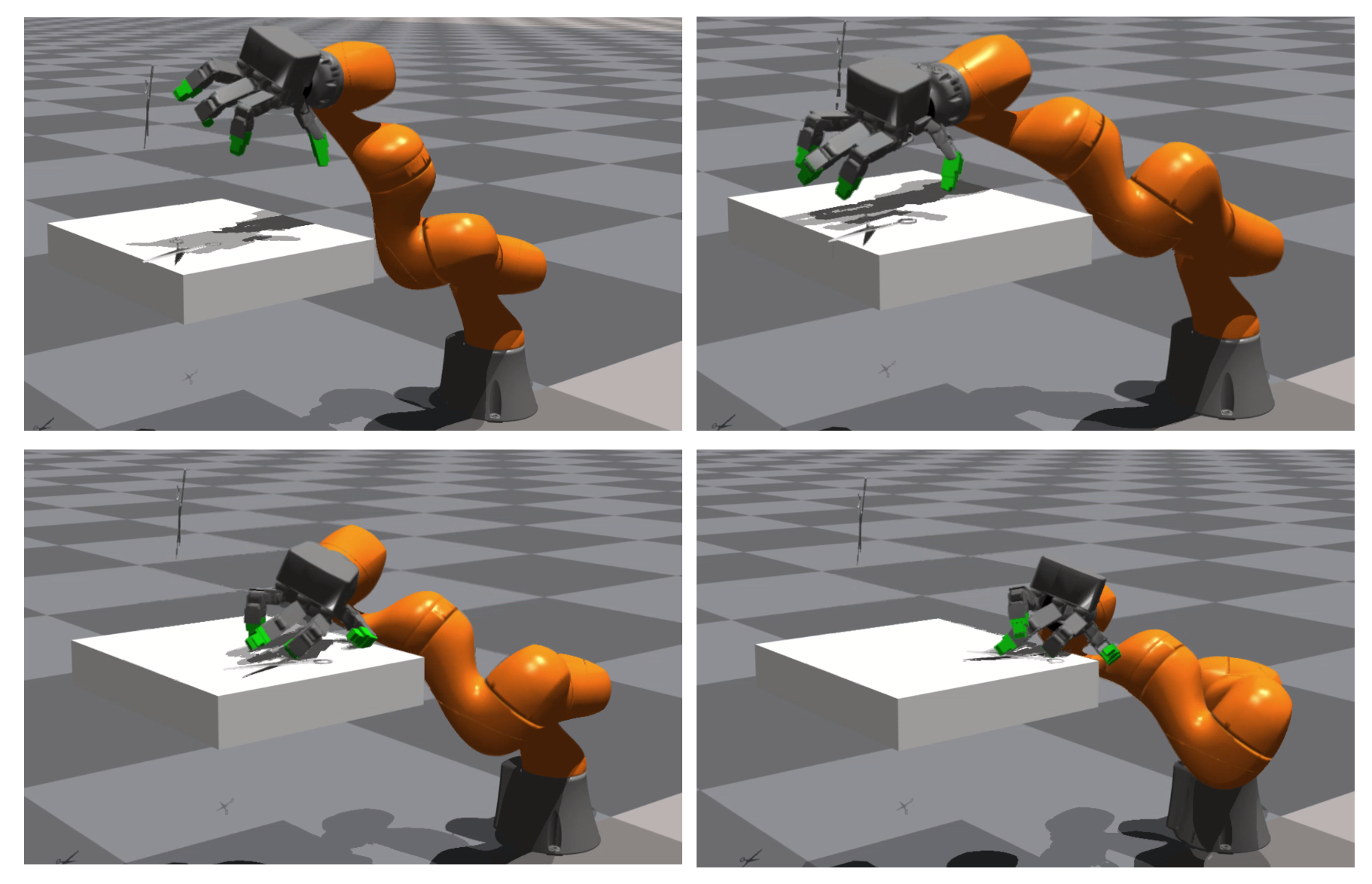}
\caption{
Using our framework and despite the absence of a specific reward indication for non-prehensile manipulation skills, the robot learns to grasp and lift the scissors by first pushing them to the edge of the table.}
\label{Fig:Figure_4}
\end{figure}

\begin{figure*}[t]
\centering
\includegraphics[width=1.0\linewidth,keepaspectratio]{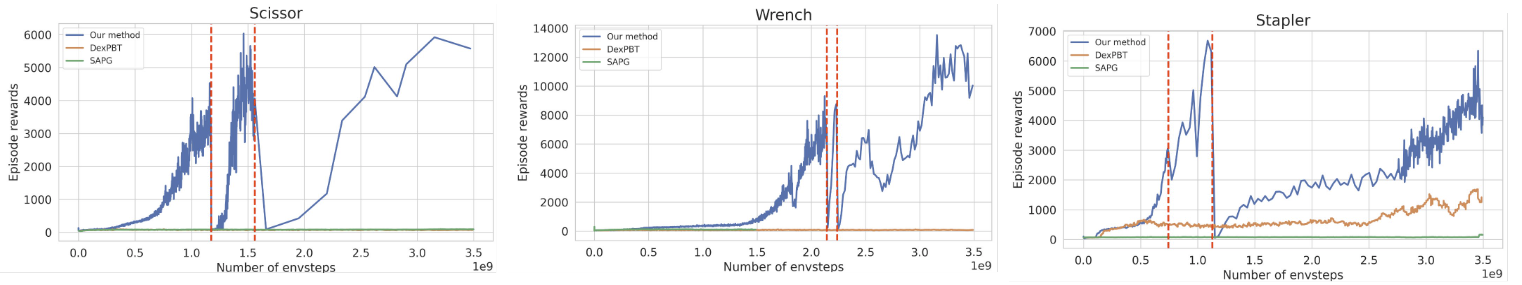}
\caption{
Reward curves comparing our method with DexPBT and SAPG on three challenging objects indicate that our framework, represented by the blue lane, performs well on all objects. In contrast, DexPBT converge to a low reward value for the relatively large object, the stapler, and fail on the other two objects. The red line indicate when the stage changes.}
\label{Fig:Figure_5}
\end{figure*}

\subsection{Dexterous manipulation of challenge YCB objects}\label{dexterous_experiment}
\textbf{Experiment setup.} 
To evaluate the generalization capability of our proposed method across different platforms, we conducted experiments involving the grasping of thin objects using a Kuka robot arm with an AllegroHand (e.g. scissors, stapler and wrench). These tasks involve grasping objects from a tabletop, lifting them, and reorienting them to achieve a specific target pose. 

The work~\cite{wu2024unidexfpm} focuses on functional grasping of various objects with Shadow hand using a carefully designed reward function, a novel learning framework, and predefined grasp poses. However, their results indicate that grasping thin objects remains a significant challenge. Therefore, to further validate our method, we selected three objects: scissors, stapler, and wrench, which this approach failed to grasp successfully.

For this experiment, we utilized the same environment setup provided by DexPBT~\cite{petrenko2023dexpbt}. DexPBT employs a population-based training method to enhance the exploration capabilities of deep reinforcement learning. Additionally, SAPG~\cite{sapg2024}, another approach utilizing the same environment as DexPBT, proposes an efficient way to leverage large-scale environments by partitioning them into smaller chunks and recombining them via importance sampling. These methods are the current SOTA for this environment setup and they were chosen as baselines for comparison with our approach. To ensure a fair comparison, all experimental setups, as well as the policy’s observations and reward, were kept identical across methods.

\textbf{Results in simulation.}  
As shown in Fig.~\ref{Fig:Figure_4}, our approach enables the robot to successfully grasp the scissors by maneuvering it to the edge of the table. Although DexPBT and SAPG demonstrate strong capabilities in achieving rapid convergence and efficient exploration for randomly sized cubes, they struggle when applied to the more challenging YCB objects. The training progress for all three YCB objects is shown in Fig.\ref{Fig:Figure_5}. The red line indicate the stage switch. Among these objects, the stapler is relatively thick and can be grasped directly; however, the rewards for DexPBT only converged at approximately 1500. This is due to the necessity of long-horizon exploration, where the robot must establish a stable grasp with an appropriate pose while simultaneously lifting and accurately adjusting orientation of the object to achieve the desired goal.

\subsection{Sim to real transfer}\label{sim2real}
To demonstrate that the policy learned with our framework does not exploit simulation-specific shortcuts and is physically reliable, we conducted real-world experiments. In this work, we apply domain randomization to improve the robustness of the learned policy, thereby facilitating better sim-to-real transfer. Specifically, we randomize object pose and shape while adding noise to observations to improve the robustness of the learned behavior.
Our focus is on grasping objects from non-graspable poses in a tabletop setting, where occlusions frequently occur, making it challenging to obtain precise pose information. Therefore, we distilled the robot's movement trajectories from simulation and transferred them to the real environment by replicating the recorded behavior.

As shown in Fig.~\ref{fig:main}, we conducted two real-world experiments. The learned behavior of pushing objects to the edge of the table adapts to various object poses. When the object's pose is unsuitable for a direct push-and-grasp action, the robot first reorients the object before grasping it from the side.

The most challenging task is the pivot grasp, where the robot learns to use the its base as support to perform the pivot grasp task. This behavior, learned in a constrained environment, enables the robot to stably maintain the pivot grasp pose and gradually adjust the object pose for a successful grasp.

These results demonstrate that the behaviors from our method are both robust and physically reliable, validating the effectiveness of our approach in real-world settings.

\begin{figure}[t]
\centering
\includegraphics[width=1.0\linewidth,keepaspectratio]{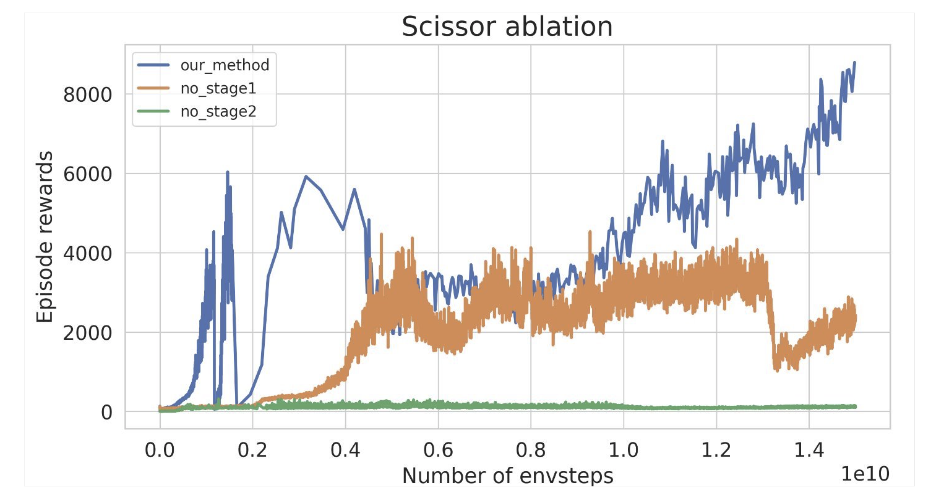}
\caption{
An ablation study on the challenging object scissors shows that with our framework, the policy quickly converges to a high reward; without stage 1, it slowly converges to a lower reward, and without stage 2, it fails to converge entirely.}
\label{Fig:Figure_7}

\end{figure}
\subsection{Ablation study}\label{ablation}
\textbf{Experiment setup.} We also performed an ablation study to evaluate the importance of the three-stage privileged action curriculum learning framework. Based on the results presented in Fig.~\ref{dexterous_experiment}, we selected the most challenging object, scissors, for this analysis.

As shown in Fig.~\ref{Fig:Figure_7}, we compared the training performance by removing specific stages from the curriculum. Specifically, we trained the policy without Stage 1 and without Stage 2 to assess their individual contributions to the overall learning process. This comparison highlights the impact of each stage on improving the policy's ability to explore and learn effectively.

\textbf{Results and analysis.}
It is obvious from Fig.~\ref{Fig:Figure_7} that without stage 2, the policy gets stuck in a local optimum. The constraint relaxation through collision management aids the robot in discovering a stable grasp strategy, which is crucial for successfully executing the subsequent in-hand orientation task. However, without the virtual force stage, the policy lacks sufficient exploration, leading to failure in finding a feasible solution for grasping the scissors under realistic collision constraints.

When training without the stage 1, the policy can still converge to a successful grasping behavior, but it requires significantly more training time. Notably, a policy trained from the virtual force stage directly can eventually learn a workable behavior after an extended period of exploration, rather than getting trapped in local optimum. This is because virtual force reduces the complexity of object-robot interactions, making it easier for the policy to explore effective interaction strategies. However, without stage 1, the policy struggles to select an optimal grasp pose, which is essential for the subsequent reorientation task, ultimately leading to a lack of long-horizon planning capability.

\section{Conclusion and Limitations}
In this work, we propose a structured framework that integrates privileged actions with curriculum learning for tackling long-horizon, contact-rich manipulation tasks. Through extensive evaluations in both simulation and real-world experiments, we demonstrate that our framework can significantly enhances the policy’s exploration efficiency. Our method not only facilitates robust non-prehensile manipulation under sparse reward conditions but also enables the robot to learn diverse behaviors within the same reward setting. Additionally, it outperforms SOTA methods on challenging dexterous manipulation tasks.

Although our approach is primarily applied to tabletop manipulation, privileged actions have the potential to extend beyond this domain and address a broader range of robot learning problems. However, a key limitation is the lack of a unified framework for optimizing privileged actions automatically across different tasks. Future work should focus on developing a generalized framework capable of dynamically adapting and optimizing privileged actions for various robotic learning challenges.

\bibliographystyle{plainnat}
\bibliography{references}

\begin{thebibliography}{40}
\providecommand{\natexlab}[1]{#1}
\providecommand{\url}[1]{\texttt{#1}}
\expandafter\ifx\csname urlstyle\endcsname\relax
  \providecommand{\doi}[1]{doi: #1}\else
  \providecommand{\doi}{doi: \begingroup \urlstyle{rm}\Url}\fi

\bibitem[Andrychowicz et~al.(2020)Andrychowicz, Baker, Chociej, Józefowicz, McGrew, Pachocki, Petron, Plappert, Powell, Ray, Schneider, Sidor, Tobin, Welinder, Weng, and Zaremba]{inhand}
OpenAI:~Marcin Andrychowicz, Bowen Baker, Maciek Chociej, Rafal Józefowicz, Bob McGrew, Jakub Pachocki, Arthur Petron, Matthias Plappert, Glenn Powell, Alex Ray, Jonas Schneider, Szymon Sidor, Josh Tobin, Peter Welinder, Lilian Weng, and Wojciech Zaremba.
\newblock Learning dexterous in-hand manipulation.
\newblock \emph{The International Journal of Robotics Research}, 39\penalty0 (1):\penalty0 3--20, 2020.
\newblock \doi{10.1177/0278364919887447}.
\newblock URL \url{https://doi.org/10.1177/0278364919887447}.

\bibitem[Atanassov et~al.(2024)Atanassov, Yu, Mitchell, Finean, and Havoutis]{atanassov2024constrained}
Vassil Atanassov, Wanming Yu, Alexander~Luis Mitchell, Mark~Nicholas Finean, and Ioannis Havoutis.
\newblock Constrained skill discovery: Quadruped locomotion with unsupervised reinforcement learning.
\newblock \emph{arXiv preprint arXiv:2410.07877}, 2024.

\bibitem[Bauza et~al.(2024)Bauza, Chen, Dalibard, Gileadi, Hafner, Martins, Moore, Pevceviciute, Laurens, Rao, et~al.]{bauza2024demostart}
Maria Bauza, Jose~Enrique Chen, Valentin Dalibard, Nimrod Gileadi, Roland Hafner, Murilo~F Martins, Joss Moore, Rugile Pevceviciute, Antoine Laurens, Dushyant Rao, et~al.
\newblock Demostart: Demonstration-led auto-curriculum applied to sim-to-real with multi-fingered robots.
\newblock \emph{arXiv preprint arXiv:2409.06613}, 2024.

\bibitem[Calli et~al.(2017)Calli, Singh, Bruce, Walsman, Konolige, Srinivasa, Abbeel, and Dollar]{calli2017yale}
Berk Calli, Arjun Singh, James Bruce, Aaron Walsman, Kurt Konolige, Siddhartha Srinivasa, Pieter Abbeel, and Aaron~M Dollar.
\newblock Yale-cmu-berkeley dataset for robotic manipulation research.
\newblock \emph{The International Journal of Robotics Research}, 36\penalty0 (3):\penalty0 261--268, 2017.

\bibitem[Chen et~al.(2021)Chen, Xu, and Agrawal]{chen2021system}
Tao Chen, Jie Xu, and Pulkit Agrawal.
\newblock A system for general in-hand object re-orientation.
\newblock \emph{Conference on Robot Learning}, 2021.

\bibitem[Chen et~al.(2023{\natexlab{a}})Chen, Tippur, Wu, Kumar, Adelson, and Agrawal]{chen2023visual}
Tao Chen, Megha Tippur, Siyang Wu, Vikash Kumar, Edward Adelson, and Pulkit Agrawal.
\newblock Visual dexterity: In-hand reorientation of novel and complex object shapes.
\newblock \emph{Science Robotics}, 8\penalty0 (84):\penalty0 eadc9244, 2023{\natexlab{a}}.

\bibitem[Chen et~al.(2023{\natexlab{b}})Chen, Wang, Fei-Fei, and Liu]{pmlr-v229-chen23e}
Yuanpei Chen, Chen Wang, Li~Fei-Fei, and Karen Liu.
\newblock Sequential dexterity: Chaining dexterous policies for long-horizon manipulation.
\newblock In Jie Tan, Marc Toussaint, and Kourosh Darvish, editors, \emph{Proceedings of The 7th Conference on Robot Learning}, volume 229 of \emph{Proceedings of Machine Learning Research}, pages 3809--3829. PMLR, 06--09 Nov 2023{\natexlab{b}}.
\newblock URL \url{https://proceedings.mlr.press/v229/chen23e.html}.

\bibitem[Chen et~al.(2024)Chen, Wang, Yang, and Liu]{chen2024object}
Yuanpei Chen, Chen Wang, Yaodong Yang, and C~Karen Liu.
\newblock Object-centric dexterous manipulation from human motion data.
\newblock \emph{arXiv preprint arXiv:2411.04005}, 2024.

\bibitem[Cheng and Xu(2023)]{cheng2023league}
Shuo Cheng and Danfei Xu.
\newblock League: Guided skill learning and abstraction for long-horizon manipulation.
\newblock \emph{IEEE Robotics and Automation Letters}, 2023.

\bibitem[Cheng et~al.(2023)Cheng, Patil, Temel, Kroemer, and Mason]{cheng2023enhancing}
Xianyi Cheng, Sarvesh Patil, Zeynep Temel, Oliver Kroemer, and Matthew~T Mason.
\newblock Enhancing dexterity in robotic manipulation via hierarchical contact exploration.
\newblock \emph{IEEE Robotics and Automation Letters}, 9\penalty0 (1):\penalty0 390--397, 2023.

\bibitem[Chiappa et~al.(2024)Chiappa, Tano, Patel, Ingster, Pouget, and Mathis]{CHIAPPA20243969}
Alberto~Silvio Chiappa, Pablo Tano, Nisheet Patel, Abigaïl Ingster, Alexandre Pouget, and Alexander Mathis.
\newblock Acquiring musculoskeletal skills with curriculum-based reinforcement learning.
\newblock \emph{Neuron}, 112\penalty0 (23):\penalty0 3969--3983.e5, 2024.
\newblock ISSN 0896-6273.
\newblock \doi{https://doi.org/10.1016/j.neuron.2024.09.002}.
\newblock URL \url{https://www.sciencedirect.com/science/article/pii/S0896627324006500}.

\bibitem[Ferrandis et~al.(2024)Ferrandis, Moura, and Vijayakumar]{ferrandis2024learning}
Juan Del~Aguila Ferrandis, Joao Moura, and Sethu Vijayakumar.
\newblock Learning visuotactile estimation and control for non-prehensile manipulation under occlusions.
\newblock In \emph{8th Annual Conference on Robot Learning}, 2024.
\newblock URL \url{https://openreview.net/forum?id=oSU7M7MK6B}.

\bibitem[Ha et~al.(2023)Ha, Florence, and Song]{ha2023scaling}
Huy Ha, Pete Florence, and Shuran Song.
\newblock Scaling up and distilling down: Language-guided robot skill acquisition.
\newblock In \emph{Conference on Robot Learning}, pages 3766--3777. PMLR, 2023.

\bibitem[Haarnoja et~al.(2024)Haarnoja, Moran, Lever, Huang, Tirumala, Humplik, Wulfmeier, Tunyasuvunakool, Siegel, Hafner, et~al.]{haarnoja2024learning}
Tuomas Haarnoja, Ben Moran, Guy Lever, Sandy~H Huang, Dhruva Tirumala, Jan Humplik, Markus Wulfmeier, Saran Tunyasuvunakool, Noah~Y Siegel, Roland Hafner, et~al.
\newblock Learning agile soccer skills for a bipedal robot with deep reinforcement learning.
\newblock \emph{Science Robotics}, 9\penalty0 (89):\penalty0 eadi8022, 2024.

\bibitem[Jiang et~al.(2023)Jiang, Chen, Cao, Bi, Lu, Zhang, Rong, and Li]{jiang2023stable}
Han Jiang, Teng Chen, Jingxuan Cao, Jian Bi, Guanglin Lu, Guoteng Zhang, Xuewen Rong, and Yibin Li.
\newblock Stable skill improvement of quadruped robot based on privileged information and curriculum guidance.
\newblock \emph{Robotics and Autonomous Systems}, 170:\penalty0 104550, 2023.

\bibitem[Kalashnikov et~al.(2022)Kalashnikov, Varley, Chebotar, Swanson, Jonschkowski, Finn, Levine, and Hausman]{kalashnikov2022scaling}
Dmitry Kalashnikov, Jake Varley, Yevgen Chebotar, Benjamin Swanson, Rico Jonschkowski, Chelsea Finn, Sergey Levine, and Karol Hausman.
\newblock Scaling up multi-task robotic reinforcement learning.
\newblock In \emph{Conference on Robot Learning}, pages 557--575. PMLR, 2022.

\bibitem[Kim et~al.(2023)Kim, Han, Kim, and Kim]{kim2023pre}
Minchan Kim, Junhyek Han, Jaehyung Kim, and Beomjoon Kim.
\newblock Pre-and post-contact policy decomposition for non-prehensile manipulation with zero-shot sim-to-real transfer.
\newblock In \emph{2023 IEEE/RSJ International Conference on Intelligent Robots and Systems (IROS)}, pages 10644--10651. IEEE, 2023.

\bibitem[Kumar et~al.(2024)Kumar, Silver, McClinton, Zhao, Proulx, Lozano-Pérez, Kaelbling, and Barry]{kumar2024practice}
Nishanth Kumar, Tom Silver, Willie McClinton, Linfeng Zhao, Stephen Proulx, Tomás Lozano-Pérez, Leslie~Pack Kaelbling, and Jennifer Barry.
\newblock Practice makes perfect: Planning to learn skill parameter policies.
\newblock In \emph{Robotics: Science and Systems (RSS)}, 2024.

\bibitem[Lee et~al.(2020)Lee, Hwangbo, Wellhausen, Koltun, and Hutter]{lee2020learning}
Joonho Lee, Jemin Hwangbo, Lorenz Wellhausen, Vladlen Koltun, and Marco Hutter.
\newblock Learning quadrupedal locomotion over challenging terrain.
\newblock \emph{Science robotics}, 5\penalty0 (47):\penalty0 eabc5986, 2020.

\bibitem[Loquercio et~al.(2021)Loquercio, Kaufmann, Ranftl, M{\"u}ller, Koltun, and Scaramuzza]{loquercio2021learning}
Antonio Loquercio, Elia Kaufmann, Ren{\'e} Ranftl, Matthias M{\"u}ller, Vladlen Koltun, and Davide Scaramuzza.
\newblock Learning high-speed flight in the wild.
\newblock \emph{Science Robotics}, 6\penalty0 (59):\penalty0 eabg5810, 2021.

\bibitem[Luo et~al.(2024)Luo, Xu, Geng, Feng, Fang, Tan, Schaal, and Levine]{luo2024multi}
Jianlan Luo, Charles Xu, Xinyang Geng, Gilbert Feng, Kuan Fang, Liam Tan, Stefan Schaal, and Sergey Levine.
\newblock Multi-stage cable routing through hierarchical imitation learning.
\newblock \emph{IEEE Transactions on Robotics}, 2024.

\bibitem[Makoviychuk et~al.(2021)Makoviychuk, Wawrzyniak, Guo, Lu, Storey, Macklin, Hoeller, Rudin, Allshire, Handa, et~al.]{makoviychuk2021isaac}
Viktor Makoviychuk, Lukasz Wawrzyniak, Yunrong Guo, Michelle Lu, Kier Storey, Miles Macklin, David Hoeller, Nikita Rudin, Arthur Allshire, Ankur Handa, et~al.
\newblock Isaac gym: High performance gpu-based physics simulation for robot learning.
\newblock \emph{arXiv preprint arXiv:2108.10470}, 2021.

\bibitem[Mao et~al.(2024)Mao, Giudici, Coppola, Althoefer, Farkhatdinov, Li, and Jamone]{10802807}
Xiaofeng Mao, Gabriele Giudici, Claudio Coppola, Kaspar Althoefer, Ildar Farkhatdinov, Zhibin Li, and Lorenzo Jamone.
\newblock Dexskills: Skill segmentation using haptic data for learning autonomous long-horizon robotic manipulation tasks.
\newblock In \emph{2024 IEEE/RSJ International Conference on Intelligent Robots and Systems (IROS)}, pages 5104--5111, 2024.
\newblock \doi{10.1109/IROS58592.2024.10802807}.

\bibitem[Margolis et~al.(2024)Margolis, Yang, Paigwar, Chen, and Agrawal]{margolis2024rapid}
Gabriel~B Margolis, Ge~Yang, Kartik Paigwar, Tao Chen, and Pulkit Agrawal.
\newblock Rapid locomotion via reinforcement learning.
\newblock \emph{The International Journal of Robotics Research}, 43\penalty0 (4):\penalty0 572--587, 2024.

\bibitem[Miki et~al.(2022)Miki, Lee, Hwangbo, Wellhausen, Koltun, and Hutter]{miki2022learning}
Takahiro Miki, Joonho Lee, Jemin Hwangbo, Lorenz Wellhausen, Vladlen Koltun, and Marco Hutter.
\newblock Learning robust perceptive locomotion for quadrupedal robots in the wild.
\newblock \emph{Science robotics}, 7\penalty0 (62):\penalty0 eabk2822, 2022.

\bibitem[Mishra et~al.(2023)Mishra, Xue, Chen, and Xu]{mishra2023generative}
Utkarsh~Aashu Mishra, Shangjie Xue, Yongxin Chen, and Danfei Xu.
\newblock Generative skill chaining: Long-horizon skill planning with diffusion models.
\newblock In \emph{Conference on Robot Learning}, pages 2905--2925. PMLR, 2023.

\bibitem[Mordatch et~al.(2012)Mordatch, Popovi{\'c}, and Todorov]{mordatch2012contact}
Igor Mordatch, Zoran Popovi{\'c}, and Emanuel Todorov.
\newblock Contact-invariant optimization for hand manipulation.
\newblock In \emph{Proceedings of the ACM SIGGRAPH/Eurographics symposium on computer animation}, pages 137--144, 2012.

\bibitem[Petrenko et~al.(2023)Petrenko, Allshire, State, Handa, and Makoviychuk]{petrenko2023dexpbt}
Aleksei Petrenko, Arthur Allshire, Gavriel State, Ankur Handa, and Viktor Makoviychuk.
\newblock Dexpbt: Scaling up dexterous manipulation for hand-arm systems with population based training.
\newblock In \emph{RSS}, 2023.

\bibitem[Qi et~al.(2023)Qi, Yi, Suresh, Lambeta, Ma, Calandra, and Malik]{qi2023general}
Haozhi Qi, Brent Yi, Sudharshan Suresh, Mike Lambeta, Yi~Ma, Roberto Calandra, and Jitendra Malik.
\newblock General in-hand object rotation with vision and touch.
\newblock In \emph{Conference on Robot Learning}, pages 2549--2564. PMLR, 2023.

\bibitem[Singla et~al.(2024)Singla, Agarwal, and Pathak]{sapg2024}
Jayesh Singla, Ananye Agarwal, and Deepak Pathak.
\newblock Sapg: Split and aggregate policy gradients.
\newblock In \emph{Proceedings of the 41st International Conference on Machine Learning (ICML 2024)}, Proceedings of Machine Learning Research, Vienna, Austria, July 2024. PMLR.

\bibitem[Triantafyllidis et~al.(2023)Triantafyllidis, Acero, Liu, and Li]{triantafyllidis2023hybrid}
Eleftherios Triantafyllidis, Fernando Acero, Zhaocheng Liu, and Zhibin Li.
\newblock Hybrid hierarchical learning for solving complex sequential tasks using the robotic manipulation network roman.
\newblock \emph{Nature Machine Intelligence}, 5\penalty0 (9):\penalty0 991--1005, 2023.

\bibitem[von Hartz et~al.(2024)von Hartz, Welschehold, Valada, and Boedecker]{von2024art}
Jan~Ole von Hartz, Tim Welschehold, Abhinav Valada, and Joschka Boedecker.
\newblock The art of imitation: Learning long-horizon manipulation tasks from few demonstrations.
\newblock \emph{IEEE Robotics and Automation Letters}, 2024.

\bibitem[Wang et~al.(2023)Wang, Fan, Sun, Zhang, Fei-Fei, Xu, Zhu, and Anandkumar]{wang2023mimicplay}
Chen Wang, Linxi Fan, Jiankai Sun, Ruohan Zhang, Li~Fei-Fei, Danfei Xu, Yuke Zhu, and Anima Anandkumar.
\newblock Mimicplay: Long-horizon imitation learning by watching human play.
\newblock In \emph{Conference on Robot Learning}, pages 201--221. PMLR, 2023.

\bibitem[Wang et~al.(2024)Wang, Liu, Chang, Huan, and Cheng]{wang2024multi}
Dexin Wang, Chunsheng Liu, Faliang Chang, Hengqiang Huan, and Kun Cheng.
\newblock Multi-stage reinforcement learning for non-prehensile manipulation.
\newblock \emph{IEEE Robotics and Automation Letters}, 2024.

\bibitem[Wu et~al.(2024)Wu, Gan, Wu, Cheng, Yang, Zhu, and Dong]{wu2024unidexfpm}
Tianhao Wu, Yunchong Gan, Mingdong Wu, Jingbo Cheng, Yaodong Yang, Yixin Zhu, and Hao Dong.
\newblock Unidexfpm: Universal dexterous functional pre-grasp manipulation via diffusion policy.
\newblock \emph{arXiv preprint arXiv:2403.12421}, 2024.

\bibitem[Yang et~al.(2024)Yang, Lu, Church, Lin, Ford, Li, Psomopoulou, Barton, and Lepora]{yang2024anyrotate}
Max Yang, Chenghua Lu, Alex Church, Yijiong Lin, Chris Ford, Haoran Li, Efi Psomopoulou, David~AW Barton, and Nathan~F Lepora.
\newblock Anyrotate: Gravity-invariant in-hand object rotation with sim-to-real touch.
\newblock \emph{arXiv preprint arXiv:2405.07391}, 2024.

\bibitem[Yu et~al.(2024)Yu, Dunion, Li, and Albrecht]{yu2024skill}
Xuehui Yu, Mhairi Dunion, Xin Li, and Stefano~V Albrecht.
\newblock Skill-aware mutual information optimisation for generalisation in reinforcement learning.
\newblock \emph{arXiv preprint arXiv:2406.04815}, 2024.

\bibitem[Zhou and Held(2023)]{zhou2023learning}
Wenxuan Zhou and David Held.
\newblock Learning to grasp the ungraspable with emergent extrinsic dexterity.
\newblock In \emph{Conference on Robot Learning}, pages 150--160. PMLR, 2023.

\bibitem[Zhou et~al.(2024)Zhou, Garg, Fox, Garrett, and Mandlekar]{zhou2024spire}
Zihan Zhou, Animesh Garg, Dieter Fox, Caelan~Reed Garrett, and Ajay Mandlekar.
\newblock {SPIRE}: Synergistic planning, imitation, and reinforcement learning for long-horizon manipulation.
\newblock In \emph{8th Annual Conference on Robot Learning}, 2024.
\newblock URL \url{https://openreview.net/forum?id=cvUXoou8iz}.

\bibitem[Zhuang et~al.(2023)Zhuang, Fu, Wang, Atkeson, Schwertfeger, Finn, and Zhao]{zhuang2023robot}
Ziwen Zhuang, Zipeng Fu, Jianren Wang, Christopher Atkeson, Soeren Schwertfeger, Chelsea Finn, and Hang Zhao.
\newblock Robot parkour learning.
\newblock \emph{arXiv preprint arXiv:2309.05665}, 2023.

\end{thebibliography}

\end{document}